\documentclass[runningheads]{llncs}
\usepackage[T1]{fontenc}
\usepackage{subcaption}
\usepackage{adjustbox}
\usepackage{graphicx}
\usepackage{amsmath}
\usepackage{amssymb}
\usepackage{booktabs}
\usepackage{multirow}
\usepackage{booktabs}
\usepackage{xspace}
\usepackage{enumitem}
\usepackage{xcolor}
\usepackage{tikz}

\definecolor{myblue}{HTML}{007FFF}
\definecolor{mygreen}{HTML}{67AB9F}
\definecolor{mypink}{HTML}{C000C0}
\newcommand{\bluecircled}[1]{%
  \begin{tikzpicture}[baseline=(char.base)]
    \node[shape=circle, draw=myblue, fill=myblue, text=white, inner sep=1pt] (char) {#1};
  \end{tikzpicture}%
}

\usepackage[colorlinks=true, linkcolor=blue, citecolor=green, urlcolor=magenta]{hyperref}

\newcommand{\modelname}{EpiSAM\xspace}

\begin{document}
\title{\modelname: Character Segmentation in Challenging Stone Inscriptions}
\titlerunning{\modelname}
%
%

\author{
Arnav Sharma\thanks{These authors contributed equally to this work.}\orcidID{0009-0001-8367-8639}
\and
Pratyush Jena$^{\ast}$\orcidID{0009-0009-7681-7977}
\and
Amal Joseph\orcidID{0009-0005-3466-5558}
\and
Ravi Kiran Sarvadevabhatla\thanks{Corresponding author.}\orcidID{0000-0003-4134-1154}
}

\authorrunning{Sharma et al.}

\institute{
Center for Visual Information Technology, International Institute of Information Technology, Hyderabad,\\
Gachibowli, Hyderabad, Telangana 500032, India\\
\email{\{arnav.sharma, pratyush.jena, amal.joseph\}@research.iiit.ac.in} \\
\email{ravi.kiran@iiit.ac.in}\\
\url{https://ihdia.iiit.ac.in/episam/}
}
\maketitle              
\begin{abstract}
Stone inscriptions are invaluable sources of historical and linguistic knowledge, yet their automated analysis remains a major challenge due to surface irregularities, erosion, and low visual contrast. Conventional document and handwriting analysis techniques fail to perform well in these scenarios. In this work, we propose character detection as a core strategy for robust inscription analysis. We introduce \textbf{\modelname}, a prompt-guided transformer framework for character segmentation in stone inscriptions. 
Rather than treating characters in isolation, \textbf{\modelname} employs a novel neighbor-aware strategy, explicitly predicting adjacent characters alongside the target. These contextual cues resolves boundary ambiguities, improving mask generation and enabling more accurate character segmentation. Furthermore, we expand an existing stone inscription dataset by adding dense polygonal annotations for characters, thereby enabling  comprehensive research on Southeast Asian epigraphy. Experimental results shows that \textbf{\modelname} achieves consistent improvements over existing baselines, while also exhibiting strong zero-shot generalization in challenging epigraphic scenarios. 

\end{abstract}


\section{Introduction}\label{sec:introduction}

Text preserved on stone surfaces such as carvings, engravings, and impressions forms a vital part of cultural and linguistic heritage. While recent advances in deep learning~\cite{barrere2022light,baena2024general,ye2024hi,fujitake2024dtrocr,Gunda_2026_WACV} have significantly improved printed and handwritten document analysis, applying similar techniques to epigraphic materials remains challenging. Stone inscriptions exhibit substantial degradation due to erosion, uneven illumination, and the irregular geometry of carved surfaces, leading to low contrast and noisy backgrounds. These factors make text detection and recognition difficult. Consequently, despite their historical importance, computational analysis of inscriptions remains severely underexplored.

\begin{figure}
    \centering
    \includegraphics[width=0.65\linewidth]{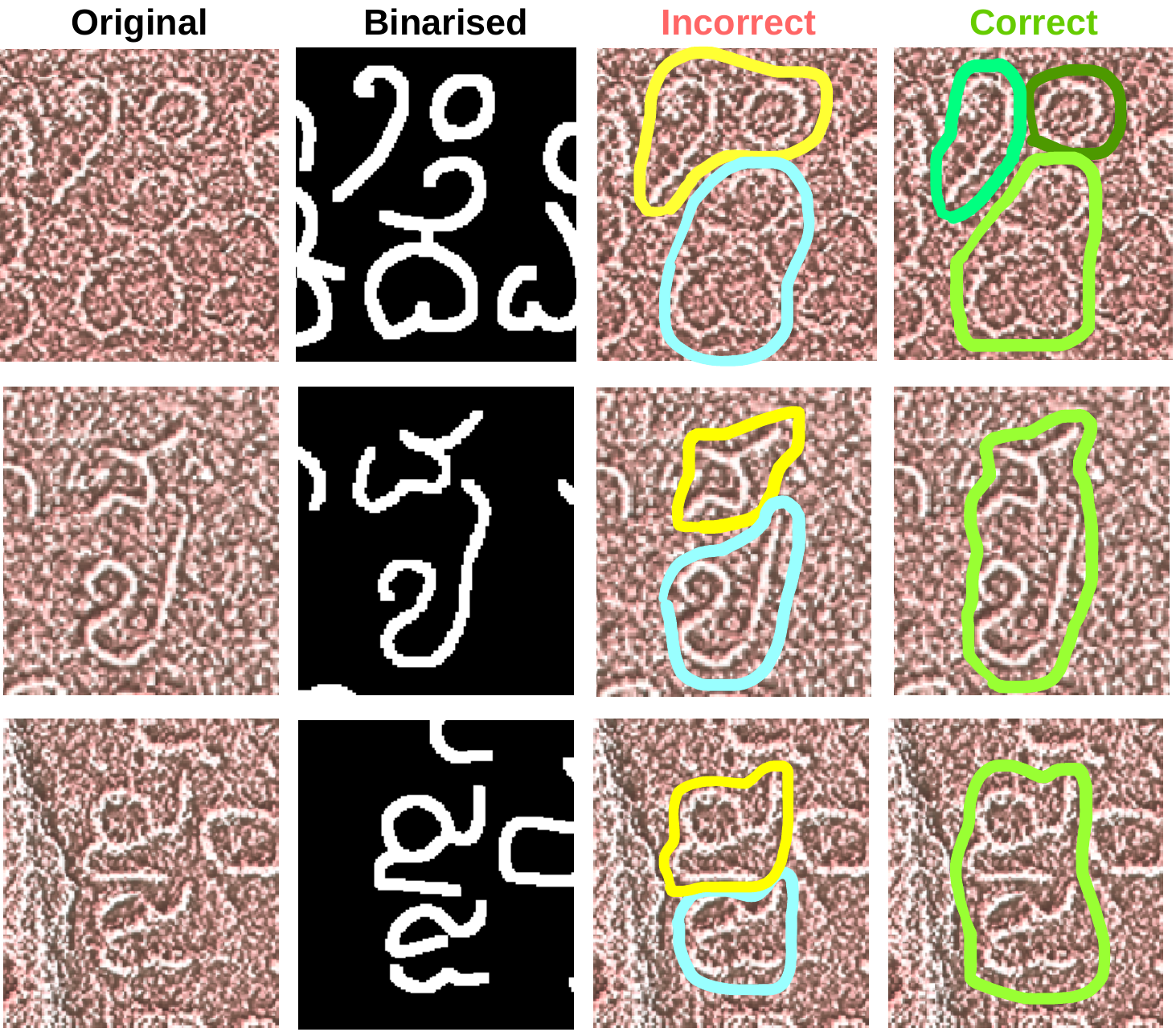}
    \caption{Notice the challenge in character isolation
due to compound glyphs and ligatures, where spatial proximity does not always indicate character
boundaries.}
    \label{fig:semantic_difficulty}
\end{figure}

Classical document systems relied on explicit character segmentation as a precursor to recognition~\cite{casey2002survey}, whereas modern frameworks often favor end-to-end recognition without detailed segmentation outputs. While effective on clean, structured text, such approaches struggle when characters are eroded, fragmented, or irregularly arranged—conditions typical of stone inscriptions. Furthermore, simple spatial proximity is often insufficient for reliably define character boundaries. In compound glyphs and ligatures, especially common in Indic scripts, components of a single character may be spatially separated, while neighboring components may belong to different glyphs (see Fig.~\ref{fig:semantic_difficulty}).
\vspace{-10pt}

\vspace{-10pt}
To address these challenges, we formulate character segmentation as a foundation for robust inscription analysis in low-resource and visually degraded settings. We propose \textbf{\modelname}, a prompt-guided transformer architecture for character segmentation. Our method leverages the strong visual priors and prompt-based localization capabilities of the Segment Anything Model (SAM)~\cite{kirillov2023segment} to produce fine-grained masks. Crucially, we move beyond isolated character prediction by introducing an adjacency prior that models the target character alongside its immediate neighbors. This neighbor-aware formulation provides contextual cues that reduce boundary ambiguities and encourage spatial consistency among adjacent characters, leading to more reliable character segmentation.

\begin{figure}[!t]
    \centering
    \includegraphics[width=\linewidth]{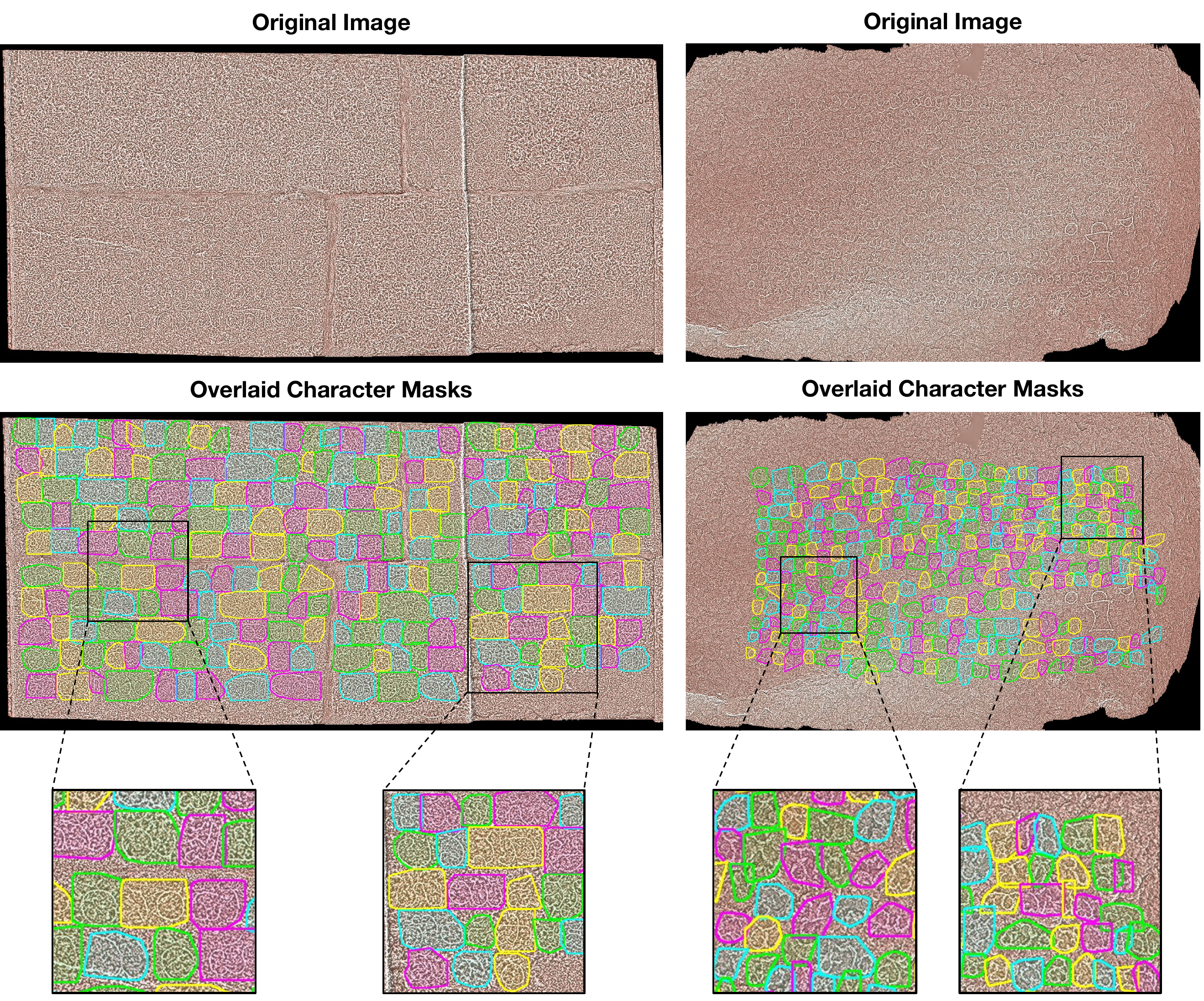}
    \caption{Sample images and their corresponding character masks from our dataset. Notice the difficulty distinguishing the shallow handwritten text etching from the background stone texture with naked eye. \textit{For clarity, the character masks are shown as boundary overlays.}}
    \label{fig:dataset_example}
\end{figure}

Unlike modern scene-text or document domains with massive corpora, inscription analysis remains severely low-resource. High-quality epigraphic data is rare, difficult to digitize, and expensive to annotate. To mitigate this scarcity and rigorously evaluate our approach, we extend an existing stone inscription dataset~\cite{jena2026unveiling} by adding precise character-level polygonal annotations. This curated dataset comprises heavily degraded artifacts spanning diverse historical periods, etching styles, and physical conditions, where foreground carvings are often visually indistinguishable from background noise (see Fig.~\textbf{\ref{fig:dataset_example}}). \\

\noindent In summary, our main contributions are:
\begin{enumerate}[label=(\roman*)]
    \item We propose \textbf{\modelname}, a prompt-guided transformer for character segmentation in severely degraded epigraphic images that incorporates a neighbor-aware strategy to resolve boundary ambiguities and enforce spatial consistency among adjacent characters.
    \item We introduce a densely annotated stone inscription dataset with precise polygonal annotations at character level.
\end{enumerate}

\section{Related Works}\label{sec:prev_works}
\subsection{Text Detection and Segmentation}\label{sec:prev_works-text_detection}
Detection and segmentation of characters and lines in documents plays a central role in text-understanding systems, bridging low-level visual processing and higher-level recognition tasks. Early research~\cite{casey2002survey,fujisawa1992segmentation,bengio1995lerec,neumann2012real,pan2010hybrid} primarily relied on handcrafted heuristics and structural analysis to isolate individual characters from printed or handwritten text. With the advent of deep learning, modern approaches~\cite{jaderberg2014deep,huang2014robust,yao2016scene,zhang2016multi} achieved significant improvements in detection robustness and generalization.

In subsequent years, the focus of text detection gradually shifted from character-level~\cite{epshtein2010detecting,neumann2012real} to word-level~\cite{zhou2017east,zhang2022arbitrary,ye2023dptext} and line-level ~\cite{wigington2018start,long2018textsnake,vadlamudi2023seamformer,monnier2020docextractor} representations to better capture contextual and layout information. Text line detection and segmentation in historical documents are also well studied~\cite{rabaev2025recent,zottin2025icdar,sterzinger2025few,hu2024seghist,agrawal2025linetr,chincholikar2025case}. More recently, there has been a renewed interest in character-level detection~\cite{baena2024general,xing2019convolutional,baek2019character,baek2020cleval}, driven by the need for fine-grained recognition and multi-script adaptability. Despite these advancements, even state-of-the-art detectors struggle in highly degraded inscription imagery, where etching marks, stone textures, and surface noise are often visually indistinguishable.

\vspace{-10pt}
\subsection{Computational Epigraphy and Inscription Analysis}\label{sec:prev_works-computational_epigraphy}
Research on image analysis and text recognition in stone inscriptions remains limited compared to other document domains. Kumar et al.~\cite{kumar2024review} provide a comprehensive review of computational methods for epigraphic documents. Recent works on automated inscription analysis has primarily focused on adopting OCR~\cite{howe2025character,bhuvaneswari2024enhancing,sekharan2025veda,agrawal2024optical}, and object detection~\cite{mohammed2024detection,zhen2024oracle,tao2025clustering,fu2024detecting,qi2025ancientglyphnet} frameworks tailored to specific ancient scripts. Deep convolutional~\cite{preethi2023region,guo2023applications} and transformer-based architectures~\cite{zheng2024ancient,murugan2025gan} have demonstrated promising results. Recent methods extend baseline visual recognition frameworks with domain-specific heuristics~\cite{zhen2024oracle} or auxiliary supervision signals~\cite{tao2025clustering,shen2025unitext} to better separate inscription characters from background noise. However, systems optimized for isolated character detection or recognition degrade significantly when characters are eroded, merged or partially missing as is the case with real-world stone inscriptions~\cite{kumar2024review,diao2025ancient}.

A critical limitation underlying many existing approaches is the scarcity and narrow scope of annotated inscription datasets~\cite{kumar2024review,mohammed2024detection}. Several works rely on synthetic augmentation pipelines~\cite{agrawal2024optical,murugan2025gan} to address this issue but fails to generalize to uncontrolled archaeological settings~\cite{diao2025ancient}. Furthermore, most publicly available datasets primarily provide character-level labels and bounding box annotations without fine-grained polygonal masks, thereby limiting research progress in precise character-level segmentation.

\vspace{-5pt}
\section{Dataset}\label{sec:dataset}

We introduce a novel dataset for character segmentation in challenging stone inscriptions, providing precise polygonal masks at character level. We extend an existing Indic stone inscription dataset~\cite{jena2026unveiling} that originally comprises of inscription images and binary text masks, by adding precise character-level bounding boxes and masks. 

The annotations were done with the help of semi-automated workflow consisting of automated template matching to get the character-wise bounding boxes followed by expert verification to ensure their correctness. The precise polygonal masks were constructed by taking the convex hull of the area inside the character bounding box.

The inscriptions span a wide range of historical periods, engraving styles and states of physical deterioration. The dataset is particularly challenging as the text and background surfaces are visually indistinguishable from the noise (See Fig. \textbf{\ref{fig:dataset_example}}). Key characteristics of the dataset are summarized in Table~\ref{tab:dataset-stats}.

\begin{table}
\caption{Dataset statistics and image properties.}
\label{tab:dataset-stats}
\centering
\begin{minipage}[t]{0.49\textwidth}
\centering
\begin{tabular}{lccc}
\hline
\multicolumn{4}{c}{\textbf{Dataset Statistics}} \\
\hline
 & Min & Max & Avg \\
\hline
Characters per image & 2 & 346 & 91 \\
\hline
\multicolumn{4}{l}{\textbf{Totals}} \\
\hline
Images & \multicolumn{3}{r}{148} \\
Characters & \multicolumn{3}{r}{13,498} \\
\hline
\end{tabular}
\end{minipage}
\hfill
\begin{minipage}[t]{0.5\textwidth}
\centering
\begin{tabular}{lccc}
\hline
\multicolumn{4}{c}{\textbf{Image Properties}} \\
\hline
 & Min & Max & Avg \\
\hline
Width (px) & 301 & 3150 & 1193 \\
Height (px) & 285 & 1947 & 1062 \\
Aspect ratio & 0.28 & 6.74 & 1.23 \\
\hline
\end{tabular}
\end{minipage}
\end{table}


\vspace{-20pt}
\section{Proposed Approach}\label{sec:approach}
\subsection{Preliminary}\label{sec:approach-preliminary}
Segment Anything (SAM) \cite{kirillov2023segment} introduced promptable image segmentation with positional cues. SAM  is composed of three modules: (a) Image encoder: a ViT-based backbone for image feature extraction; (b) Prompt encoder: encodes sparse prompts (points, boxes, text) and dense inputs (masks); (c) Mask decoder: a lightweight two-layer transformer that combines image and prompt embeddings with output tokens to predict segmentation masks and corresponding IoU scores. The output tokens act as mask tokens, analogous to the object queries in DETR~\cite{carion2020end}, and are used to predict multiple segmentation masks simultaneously. Each decoder block applies self-attention over prompt tokens and bidirectional cross-attention between prompt and image embeddings. The image features are upsampled, and each output token is passed through an MLP to produce a channel-aligned vector whose point-wise dot product with the upsampled embeddings generates the mask logits. Although SAM exhibits strong zero-shot generalization for generic segmentation, its performance falter when segmenting text in severely degraded stone inscriptions.

\subsection{Ours: \modelname}

Due to the unconstrained nature of historical epigraphy, characters often suffer from severe surface degradation, background noise, and heavily eroded boundaries. Adjacent characters are often densely interleaved, making precise instance-level segmentation particularly challenging for standard segmentation models.

Rather than predicting each character in isolation, we exploit the structured nature of inscriptions, where characters appear in locally coherent spatial sequences. We reformulate character segmentation as a localized, neighbor-aware prediction task: estimating a target character jointly with its immediate left and right neighbors (Sec.~\ref{ref:char_module}). 

This formulation introduces strong contextual constraints that reduce boundary ambiguities and improve separation between adjacent characters, especially under heavy noise and texture ambiguity. By leveraging local spatial continuity, the model achieves more reliable instance-level character segmentation without requiring any explicit line-level modeling.

Based on this formulation, we propose \textbf{\modelname}, a prompt-guided transformer framework that integrates neighbor-aware contextual awareness for robust character segmentation. An overview of the architecture is given in Fig.~\ref{fig:pipeline_a}.

\vspace{-10pt}
\subsubsection{Neighbor-Aware Character Segmentation \\}
\label{ref:char_module}
The character segmentation module of \modelname comprises four components: a prompt generator, a prompt encoder, an image encoder, and a neighbor-aware decoder.

To model local context explicitly, we construct supervision for neighboring characters by computing the convex hull of the first two adjacent characters on the left and right of the target character. This provides spatially coherent neighbor masks that capture contextual structure (see Fig.~3).

\begin{figure}[t]
    \centering
    \includegraphics[width=\linewidth]{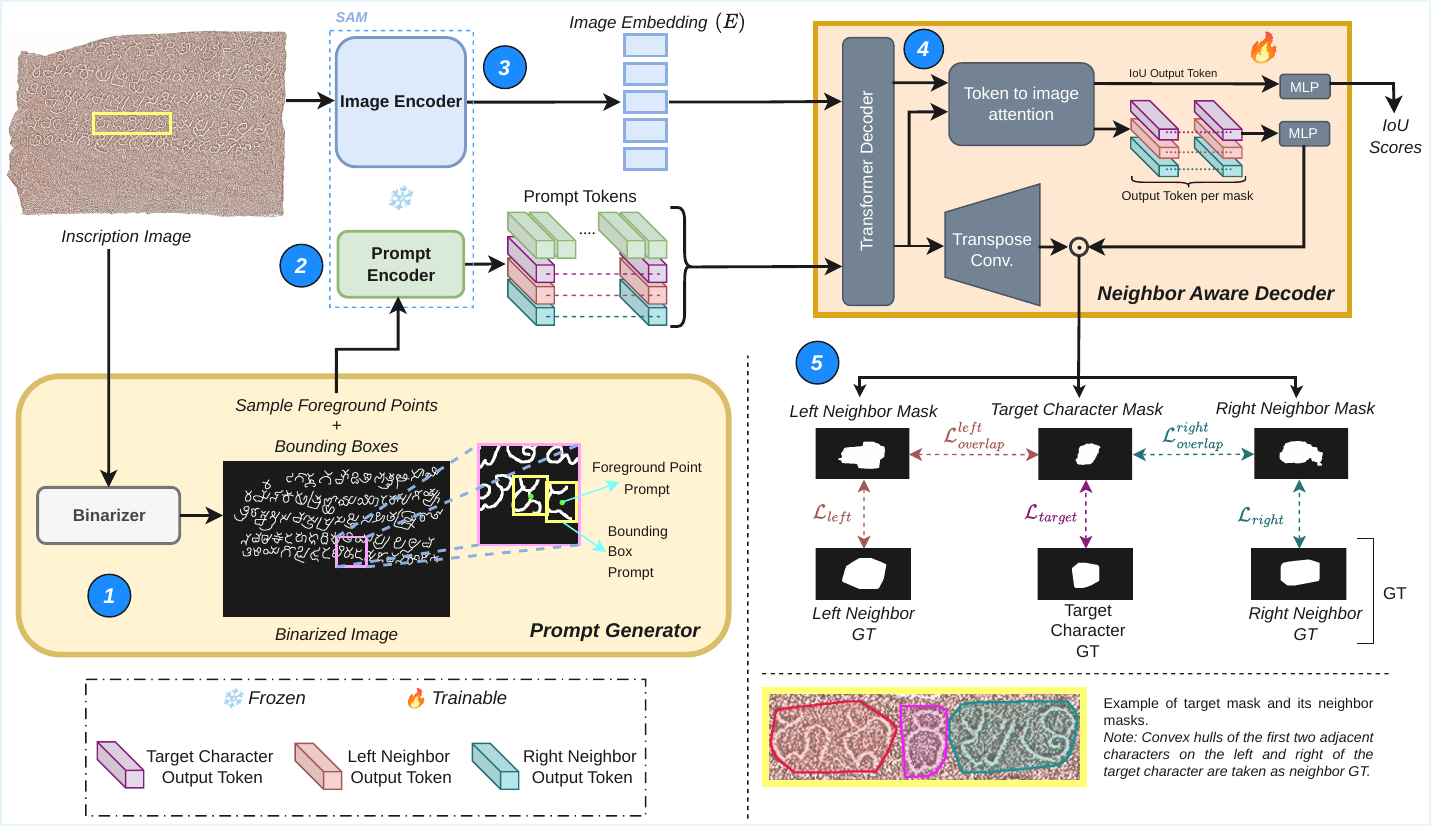}
    \caption{
        \textbf{Neighbor Aware Character Segmentation:} 
        The inscription image is \protect\bluecircled{1} binarized to generate foreground point and bounding-box prompts, which are then 
        \protect\bluecircled{2} encoded and combined with 
        \protect\bluecircled{3} frozen SAM image features. 
        (b) \protect\bluecircled{4} A neighbor-aware decoder 
        \protect\bluecircled{5} predicts the target character mask along with its left and right neighbor masks.
    }
    \label{fig:pipeline_a}
\end{figure}

\begin{enumerate}[label=(\alph*)]
    \item \textbf{Prompt Generator:} (\bluecircled{1} from Fig.~\ref{fig:pipeline_a}) We first obtain a binary foreground mask of the inscription using an Attention-UNet binarizer~\cite{jena2026unveiling}. Connected components extracted from this mask provide centroids and bounding boxes, which serve as sparse geometric prompts.
    
    \item \textbf{Prompt Encoder:} \bluecircled{2} Following SAM~\cite{kirillov2023segment}, point prompts are encoded by summing positional encodings with learned embeddings indicating foreground or background. A bounding box is represented by two corner embeddings (top-left and bottom-right), each formed by combining positional encodings with learned corner-type embeddings.
    
    \item \textbf{Image Encoder:} \bluecircled{3} Dense visual features are extracted from the RGB inscription image using SAM’s pre-trained image encoder, which remains frozen during training to leverage its robust generic representations and prevent overfitting on limited epigraphic data.
    
    \item \textbf{Neighbor-Aware Decoder:} \bluecircled{4} We extend SAM's mask decoder by introducing two additional output tokens dedicated to predicting the left and right neighbor masks. For each prompt, the decoder jointly predicts the target character mask along with it's left and right neighbors. To address boundary ambiguities, we retain SAM's original three-token structure for the target character mask, as detailed in Sec.~\ref{sec:inference}
\end{enumerate}
\noindent
\bluecircled{5}We design a loss function that encourages precise character boundary learning while enforcing contextual awareness of neighboring characters. Loss function is explained in detail in Sec.~\ref{sec:loss_function}

\vspace{-5pt}
\subsection{Prompt-Guided, Neighbor-Aware Character Segmentation}\label{sec:approach-character_segmentation}
Each training sample consists of an inscription image which is resized and padded to $I \in \mathbb{R}^{3 \times 1024 \times 1024}$, a set of foreground point prompts $\{ p_i \}_{i=1}^N$ and their corresponding box prompts $\{ b_i \}_{i=1}^N$ for each character component  $C_i$ in the binary mask \bluecircled{1}. For each character component $C_i$, the training data also includes its ground truth mask $M_i$, as well as the ground truth masks for its immediate left $(M_i^L)$ and right $(M_i^R)$ neighbors. Note that multiple connected components may correspond  to the same ground-truth character mask. 

Here, $N$ denotes the total number of character components in the inscription image. For each $C_i$, a single foreground point prompt is sampled at the center of its bounding box. The coordinates of the point prompts $(p_i)$ and box prompts $(b_i)$ are transformed using the same resizing and padding operations applied to the input image to maintain spatial alignment.

\noindent

\bluecircled{2} Each foreground point--box pair $(p_i, b_i)$ is treated as a positive prompt and encoded by the prompt encoder into sparse prompt tokens, comprising one token for the point and two tokens for the box corners, collectively denoted as
\[
T_i^{\text{prompt}} = \left( p_i', b_{i,1}', b_{i,2}' \right).
\]

\noindent
\bluecircled{3} The inscription image $I$ is processed by the image encoder $f_{\theta_{\text{enc}}}$ to produce image embeddings
\[
E = f_{\theta_{\text{enc}}}(I).
\]

\noindent
\bluecircled{4} The mask decoder $f_{\theta_{\text{dec}}}$ takes image embedding $E$ and prompt tokens $T_i^{\text{prompt}}$ as input and generates candidate character-level segmentation masks for each character $C_i$.

To address boundary ambiguities, SAM predicts three candidate masks per prompt. \bluecircled{5} We extend the decoder to produce two additional masks corresponding to the left and right neighboring characters, enabling neighbor-aware segmentation. The decoder also predicts an IoU score for each of the five output masks.

\[
\left( \{P_i^{(k)}\}_{k=1}^{5}, \{\widehat{\text{IoU}}_i^{(k)}\}_{k=1}^{5} \right)
=
f_{\theta_{\text{dec}}}(E, T_i^{\text{prompt}})
\]
, where $\{P_i^{(k)}\}_{k=1}^{5}$ and $\{\widehat{\text{IoU}}_i^{(k)}\}_{k=1}^{5}$ are the five output masks and predicted IoU scores respectively.

\vspace{-10pt}
\subsection{Neighbor-Aware Loss Function}\label{sec:loss_function}
We extend the Focal + Dice loss formulation of SAM~\cite{kirillov2023segment} to jointly supervise all five mask predictions (three target masks and two neighbor masks). The overall loss is defined as:

\[
\mathcal{L}_{total} = \mathcal{L}_{target} + \mathcal{L}_{left} + \mathcal{L}_{right} + \mathcal{L}_{overlap} + \mathcal{L}_{IoU}
\]

\paragraph{Target Mask Loss.}
From the three candidate predictions for the target character, $\{P_i^{(k)}\}_{k=1}^{3}$, we select the best mask based on a weighted combination of Focal and Dice losses against the ground-truth mask $M_i$:
\[
\mathcal{L}_{\text{target}} =
\min_{k \in \{1,2,3\}}
\left(
\lambda_{\text{focal}} \,
\mathcal{L}_{\text{focal}}(P_i^{(k)}, M_i)
+
\mathcal{L}_{\text{dice}}(P_i^{(k)}, M_i)
\right).
\]

\paragraph{Neighbor Losses.}
The neighbor losses $\mathcal{L}_{\text{left}}$ and
$\mathcal{L}_{\text{right}}$ are computed similarly using the neighbor mask predictions, $P_i^{(4)}$ and $P_i^{(5)}$, supervised against their respective ground-truth masks $M_i^{L}$ and $M_i^{R}$.

\paragraph{Overlap Loss.}
To encourage spatially distinct predictions, we introduce overlap penalties between the target mask and its neighbors, as well as between the target mask and non-associated text lines.

Let $P_i^{(k^*)}$ denote the selected best target mask. The character-level overlap penalty is defined as:

\[
\mathcal{L}_{\text{overlap}} =
\frac{1}{2}
\left(
\mathbb{E}\big[\sigma(P_i^{(k^*)}) \cdot
\sigma(P_i^{(4)})\big]
+
\mathbb{E}\big[\sigma(P_i^{(k^*)}) \cdot
\sigma(P_i^{(5)})\big]
\right),
\]
where $\sigma(\cdot)$ denotes the sigmoid function and $\mathbb{E}[\cdot]$ denotes pixel-wise averaging.

\paragraph{IoU Prediction Loss.}
The IoU regression loss supervises the five predicted IoU scores $\{\widehat{\mathrm{IoU}}_i^{(k)}\}_{k=1}^{5}$ using an L1 loss against the
true IoU computed between each predicted mask and its corresponding ground-truth mask as done in SAM 2~\cite{ravi2024sam}:

During training, ground-truth masks are resized and padded to $1024 \times 1024$ to match the transformed input images, and then
downsampled to $256 \times 256$ using bilinear interpolation to align with the mask decoder output resolution.

\vspace{-5pt}
\subsection{Inference}\label{sec:inference}
\vspace{-5pt}
The inscription image is first processed by the image encoder and the binarizer to obtain image embeddings and a binary text mask. Connected component analysis on the binary mask yields candidate character components. For each component, its centroid and bounding box are provided as prompts to the prompt encoder, producing sparse prompt tokens.
The mask decoder takes the image embeddings and prompt tokens as input and predicts three candidate target masks along with two neighbor masks. The final character mask is selected as the candidate with the highest predicted IoU score. To remove redundant predictions arising from overlapping components, Non-Maximum Suppression (NMS) is applied over the predicted character masks.

\section{Implementation Details}\label{sec:implementation_details}

For fine-tuning SAM, we keep the image and prompt encoders frozen and train only the transformer decoder. Following~\cite{kirillov2023segment}, we use a linear combination of focal loss and dice loss with a ratio of 20:1 $(\lambda_{focal}=20)$. We train using AdamW optimizer with a cosine-annealing learning-rate schedule and select the best model based on the validation IoU. All experiments are conducted on one NVIDIA A6000 GPU for 100 epochs with a batch size of 1.

\section{Experiments}\label{experiments}
We evaluate our method on per-character IoU and dice coefficient (Dice) metrics. We compare our model against baselines for character segmentation and conduct ablation studies to assess the contribution of each component of our pipeline.

\vspace{-10pt}
\subsection{Results}\label{experiments-results}

We evaluate the proposed \modelname by comparing it against recent state-of-the-art character segmentation approaches. Specifically, we benchmark against CRAFT~\cite{baek2019character}, HiSAM~\cite{ye2024hi}, YOLOv8-Seg~\cite{yolov8_ultralytics}, YOLOv11-Seg~\cite{yolo11_ultralytics}, and YOLOv12-Seg~\cite{tian2025yolov12} (see Tab.~\ref{tab:segmentation_results}). We additionally evaluate the SAM-H model without fine-tuning to assess its zero-shot performance on inscriptions.

Our method achieves consistently higher IoU and Dice scores compared to both region-based approaches (CRAFT, HiSAM) and anchor-free YOLO-based segmentation models, demonstrating its robustness to irregular character shapes, surface erosion, and heavy background noise. We used official codebases for all baselines (except CRAFT, which has no official open-source code) and followed their documented preprocessing and training protocols to ensure fair comparison.

\begin{figure}[!h]
    \centering
    \includegraphics[height=0.8\textheight]{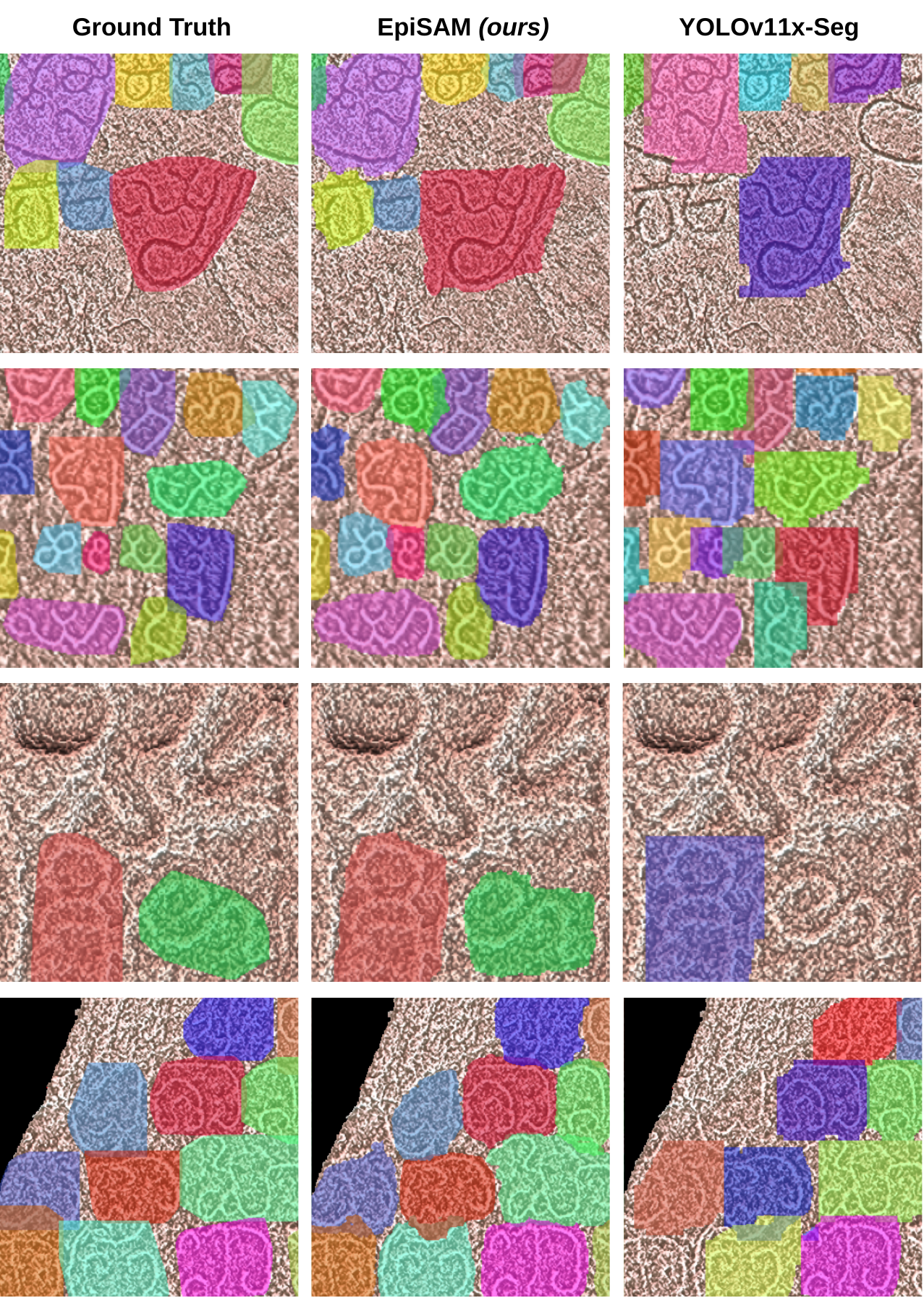}
\caption{\textbf{Qualitative comparison of character segmentation against the strongest baseline (YOLOv11x-Seg).} Under severe degradation and surface erosion, \modelname\ produces tighter and more precise character masks, resulting in improved instance separation in densely packed inscriptions.}
    \label{fig:results-qualitative_char}
    
\end{figure}

As shown in Fig.~\ref{fig:results-qualitative_char}, \modelname produces cleaner instance boundaries with significantly fewer false positives than competing baselines. The point-guided prompting mechanism improves instance separation in densely packed character clusters, while the transformer decoder enhances contextual reasoning, leading to more spatially coherent and precise character masks


\begin{table}[t]
\centering
\caption{\textbf{Quantitative results for character segmentation.} 
\textbf{\modelname} demonstrates superior instance-level segmentation performance across metrics. 
Note that standard detection methods (e.g., CRAFT) are adapted for mask-based segmentation evaluation.}
\label{tab:segmentation_results}
\begin{tabular}{lcc}
\toprule
\textbf{Method} 
& \textbf{Per-character IoU} $\uparrow$ 
& \textbf{Per-character Dice} $\uparrow$ \\
\midrule
CRAFT~\cite{baek2019character} & 32.67 & 44.74 \\
HiSAM~\cite{ye2024hi} & 1.88 & 3.47 \\
YOLOv8x-Seg~\cite{yolov8_ultralytics} & 52.72 & 64.24 \\
YOLOv11x-Seg~\cite{yolo11_ultralytics} & 54.74 & 67.18 \\
YOLOv12x-Seg~\cite{tian2025yolov12} & 53.75 & 65.34 \\
SAM-H (vanilla)~\cite{kirillov2023segment} & 59.99 & 72.68 \\
\midrule
\textbf{\modelname\ (ours)} & \textbf{66.46} & \textbf{78.68} \\
\bottomrule
\end{tabular}
\end{table}

\subsection{Ablation Studies}\label{experiments-ablation_studies}

To analyze the contribution of individual design choices in \modelname, we conduct several ablation experiments on the validation set of our dataset. Note that ablations~\ref{sec:ablation-prompt} and ~\ref{sec:ablation-neighbor} was performed with ViT-B as the transformer backbone.

\vspace{-10pt}
\subsubsection{Ablation on Prompting Strategy}
\label{sec:ablation-prompt}
\begin{table*}[tb!]
\centering
\caption{\textbf{Ablation on Prompting Strategies.} Comparison between centroid-only prompting and centroid-plus-bounding-box prompting, highlighting the impact of incorporating spatial extent information.}
\label{tab:ablation-prompting}
\renewcommand{\arraystretch}{1.5}

\begin{tabular*}{\textwidth}{@{\extracolsep{\fill}}lccc}
\toprule
\textbf{Prompting Method} & \textbf{Per-character IoU} $\uparrow$ & \textbf{Per-character Dice} $\uparrow$ \\
\midrule
Centroid only & 57.34  & 70.59  \\
Centroid + Bounding Box (ours) & 65.98  & 78.31 \\
\bottomrule
\end{tabular*}
\end{table*}

Table~\ref{tab:ablation-prompting} compares the prompting strategies evaluated in our study. We consider two variants: (1) using only centroid points of connected components extracted from the binary mask, and (2) using both centroid points and their corresponding bounding boxes. Incorporating bounding boxes alongside centroids in the prompts improves performance(+8.64 per-char IoU, +7.72 per-char Dice), as the additional spatial context provides stronger supervision to the model. 

\vspace{-10pt}
\subsubsection{Ablation on Neighbor Prediction}
\label{sec:ablation-neighbor}
\begin{table*}[tb!]
\centering
\caption{\textbf{Ablation on Neighbor Prediction.} Comparison between target-only and joint target–neighbor prediction, highlighting the impact of modeling adjacent character context.}
\label{tab:ablation-neighbor}
\renewcommand{\arraystretch}{1.5}

\begin{tabular*}{\textwidth}{@{\extracolsep{\fill}}lccc}
\toprule
\textbf{Prediction Setting} & \textbf{Per-character IoU} $\uparrow$ & \textbf{Per-character Dice} $\uparrow$ \\
\midrule
Target character only & 59.99 & 72.68 \\
Target + neighboring characters (ours)& 65.98 & 78.31 \\
\bottomrule
\end{tabular*}
\end{table*}

Table~\ref{tab:ablation-neighbor} compares predicting only the target character with jointly predicting the target and its neighboring characters. Incorporating neighbor-aware prediction significantly improves performance(+5.99 per-char IoU, +5.63 per-char Dice), supporting the design of our loss function. Since the model is trained to predict the neighbors along with the target character, it is able to reduce boundary ambiguity in the mask predictions.

\vspace{-10pt}
\subsubsection{Ablation on Vision Transformer Backbone \\}

\begin{table*}[tb!]
\centering
\caption{\textbf{Ablation on Vision Transformer Backbone.} Comparison of performance on different ViT backbones.}
\label{tab:ablation-vit_size}
\renewcommand{\arraystretch}{1.5}
\begin{adjustbox}{width=\linewidth}
\begin{tabular}{lcccc}
\toprule
\textbf{Backbone} & \textbf{Params (M)} & \textbf{Per-char IoU} $\uparrow$ & \textbf{Per-char Dice} $\uparrow$ & \textbf{Infer. Speed (sec/img)} $\downarrow$ \\
\midrule
ViT-B & $\sim$86M  & 65.98 & 78.31 & 0.367 \\
ViT-L & $\sim$307M & 66.46 & 78.68 & 0.526 \\
\bottomrule
\end{tabular}
\end{adjustbox}
\end{table*}

Table~\ref{tab:ablation-vit_size} compares ViT-B and ViT-L backbones~\cite{dosovitskiy2020image} for character segmentation in terms of performance, inference speed, and model size. ViT-L ($\sim$307M parameters) consistently outperforms ViT-B ($\sim$86M parameters) across all evaluation metrics, demonstrating the benefit of increased model capacity for this task. Although ViT-L incurs higher computational cost, its superior performance suggests that the larger backbone better captures fine-grained character features, even in our low-resource setting.

\vspace{-10pt}
\subsection{Qualitative Analysis}\label{experiments-qualitative_and_zeroshot}

\begin{figure}[!th]
    \centering
    \includegraphics[height=0.65\textheight]{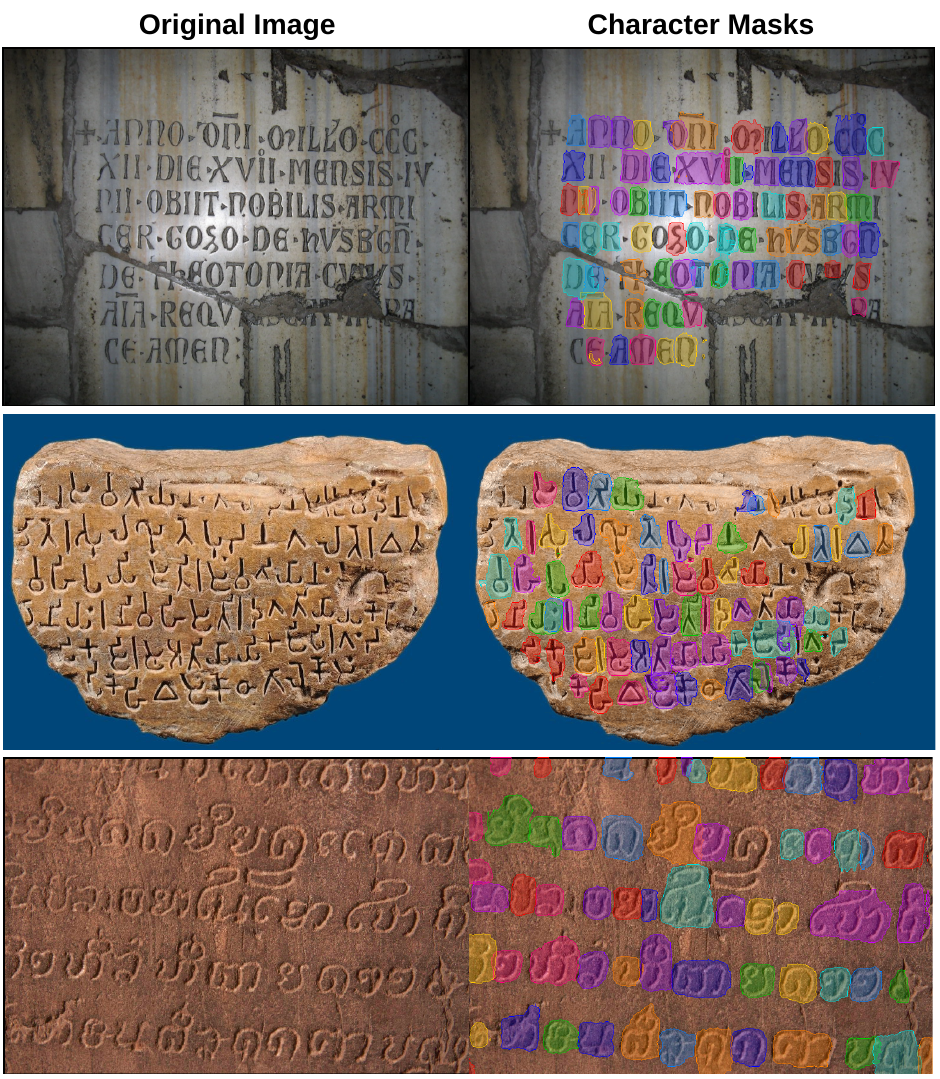}
    \caption{Zero-shot character segmentation results of \modelname\ on (from top to bottom) Roman, Brahmi and Thai inscriptions. Even when characters are partially missing or severely eroded, the model produces spatially coherent and well-separated character masks.}
    \label{fig:zero_shot}
\end{figure}

As illustrated in Fig.~\ref{fig:results-qualitative_char}, \modelname\ produces more accurate and spatially precise character masks compared to baseline methods. Competing approaches often struggle with dense, closely spaced characters and large scale variations, frequently producing merged instances or loose boundaries. In contrast, \modelname\ consistently generates well-separated and tightly aligned character masks.

For heavily degraded inscriptions, tighter instance masks are particularly beneficial for downstream tasks such as text recognition, where overly loose masks may introduce background noise and degrade recognition accuracy. This qualitative improvement is consistent with the superior per-character IoU and Dice scores reported in Sec.~\ref{experiments-results}.

Our approach demonstrates good zero-shot generalization to inscriptions other unseen scripts, spanning both Indic and non-Indic writing systems (see Fig.~\ref{fig:zero_shot}).

\section{Conclusion}\label{sec:conclusion}

In this work, we address the problem of instance-level character segmentation in severely degraded stone inscriptions. We introduce \modelname, a prompt-guided transformer framework that explicitly incorporates neighboring character context during prediction. By leveraging local adjacency cues, the proposed approach reduces boundary ambiguities and improves spatial consistency among densely packed characters. We further extend an existing inscription dataset with precise polygonal character annotations, supporting further research in digital epigraphy. Experimental results demonstrate superior segmentation accuracy compared to baselines, along with promising zero-shot generalization across diverse scripts and inscription styles.

More broadly, advances scalable computational tools for the analysis of South Asian textual heritage. Future directions include extending \modelname\ toward integrated text recognition and language modeling for end-to-end inscription transcription.

\begin{credits}
\subsubsection*{\ackname}
We acknowledge The Mythic Society Bengaluru for providing the dataset images used in this research. We gratefully acknowledge the TCS Foundation’s support for Amal Joseph under the TCS Research Scholar Program.

\end{credits}
%
\bibliographystyle{unsrt}
\bibliography{references}

\begin{thebibliography}{10}

\bibitem{barrere2022light}
Killian Barrere, Yann Soullard, Aur{\'e}lie Lemaitre, and Bertrand Co{\"u}asnon.
\newblock A light transformer-based architecture for handwritten text recognition.
\newblock In {\em International Workshop on Document Analysis Systems}, pages 275--290. Springer, 2022.

\bibitem{baena2024general}
Raphael Baena, Syrine Kalleli, and Mathieu Aubry.
\newblock General detection-based text line recognition.
\newblock {\em Neurips}, 37:42388--42404, 2024.

\bibitem{ye2024hi}
Maoyuan Ye, Jing Zhang, Juhua Liu, Chenyu Liu, Baocai Yin, Cong Liu, Bo~Du, and Dacheng Tao.
\newblock Hi-sam: Marrying segment anything model for hierarchical text segmentation.
\newblock {\em IEEE TPAMI}, 2024.

\bibitem{fujitake2024dtrocr}
Masato Fujitake.
\newblock Dtrocr: Decoder-only transformer for optical character recognition.
\newblock In {\em WACV}, pages 8025--8035, 2024.

\bibitem{Gunda_2026_WACV}
Sai~Madhusudan Gunda, Tathagata Ghosh, Simran~Singh Sandral, and Ravi~Kiran Sarvadevabhatla.
\newblock Curio: Curvature-aligned and efficient ocr for low-resource historical manuscripts.
\newblock In {\em Proceedings of the IEEE/CVF WACV}, pages 2011--2021, March 2026.

\bibitem{casey2002survey}
Richard~G Casey and Eric Lecolinet.
\newblock A survey of methods and strategies in character segmentation.
\newblock {\em IEEE TPAMI}, 18(7):690--706, 2002.

\bibitem{kirillov2023segment}
Alexander Kirillov, Eric Mintun, Nikhila Ravi, Hanzi Mao, Chloe Rolland, Laura Gustafson, Tete Xiao, Spencer Whitehead, Alexander~C Berg, Wan-Yen Lo, et~al.
\newblock Segment anything.
\newblock In {\em ICCV}, pages 4015--4026, 2023.

\bibitem{jena2026unveiling}
Pratyush Jena, Amal Joseph, Arnav Sharma, and Ravi~Kiran Sarvadevabhatla.
\newblock Unveiling text in challenging stone inscriptions: A character-context-aware patching strategy for binarization.
\newblock {\em arXiv preprint arXiv:2601.03609}, 2026.

\bibitem{fujisawa1992segmentation}
Hiromichi Fujisawa, Yasuaki Nakano, and Kiyomichi Kurino.
\newblock Segmentation methods for character recognition: from segmentation to document structure analysis.
\newblock {\em Proceedings of the IEEE}, 80(7):1079--1092, 1992.

\bibitem{bengio1995lerec}
Yoshua Bengio, Yann LeCun, Craig Nohl, and Chris Burges.
\newblock Lerec: A nn/hmm hybrid for on-line handwriting recognition.
\newblock {\em Neural computation}, 7(6):1289--1303, 1995.

\bibitem{neumann2012real}
Luk{\'a}{\v{s}} Neumann and Ji{\v{r}}{\'\i} Matas.
\newblock Real-time scene text localization and recognition.
\newblock In {\em 2012 IEEE conference on computer vision and pattern recognition}, pages 3538--3545. IEEE, 2012.

\bibitem{pan2010hybrid}
Yi-Feng Pan, Xinwen Hou, and Cheng-Lin Liu.
\newblock A hybrid approach to detect and localize texts in natural scene images.
\newblock {\em IEEE transactions on image processing}, 20(3):800--813, 2010.

\bibitem{jaderberg2014deep}
Max Jaderberg, Andrea Vedaldi, and Andrew Zisserman.
\newblock Deep features for text spotting.
\newblock In {\em European conference on computer vision}, pages 512--528. Springer, 2014.

\bibitem{huang2014robust}
Weilin Huang, Yu~Qiao, and Xiaoou Tang.
\newblock Robust scene text detection with convolution neural network induced mser trees.
\newblock In {\em European conference on computer vision}, pages 497--511. Springer, 2014.

\bibitem{yao2016scene}
Cong Yao, Xiang Bai, Nong Sang, Xinyu Zhou, Shuchang Zhou, and Zhimin Cao.
\newblock Scene text detection via holistic, multi-channel prediction.
\newblock {\em arXiv preprint arXiv:1606.09002}, 2016.

\bibitem{zhang2016multi}
Zheng Zhang, Chengquan Zhang, Wei Shen, Cong Yao, Wenyu Liu, and Xiang Bai.
\newblock Multi-oriented text detection with fully convolutional networks.
\newblock In {\em Proceedings of the IEEE CVPR}, pages 4159--4167, 2016.

\bibitem{epshtein2010detecting}
Boris Epshtein, Eyal Ofek, and Yonatan Wexler.
\newblock Detecting text in natural scenes with stroke width transform.
\newblock In {\em 2010 IEEE computer society conference on computer vision and pattern recognition}, pages 2963--2970. IEEE, 2010.

\bibitem{zhou2017east}
Xinyu Zhou, Cong Yao, He~Wen, Yuzhi Wang, Shuchang Zhou, Weiran He, and Jiajun Liang.
\newblock East: an efficient and accurate scene text detector.
\newblock In {\em Proceedings of the IEEE conference on Computer Vision and Pattern Recognition}, pages 5551--5560, 2017.

\bibitem{zhang2022arbitrary}
Shi-Xue Zhang, Xiaobin Zhu, Lei Chen, Jie-Bo Hou, and Xu-Cheng Yin.
\newblock Arbitrary shape text detection via segmentation with probability maps.
\newblock {\em IEEE transactions on pattern analysis and machine intelligence}, 45(3):2736--2750, 2022.

\bibitem{ye2023dptext}
Maoyuan Ye, Jing Zhang, Shanshan Zhao, Juhua Liu, Bo~Du, and Dacheng Tao.
\newblock Dptext-detr: Towards better scene text detection with dynamic points in transformer.
\newblock In {\em Proceedings of the AAAI conference on artificial intelligence}, volume~37, pages 3241--3249, 2023.

\bibitem{wigington2018start}
Curtis Wigington, Chris Tensmeyer, Brian Davis, William Barrett, Brian Price, and Scott Cohen.
\newblock Start, follow, read: End-to-end full-page handwriting recognition.
\newblock In {\em Proceedings of the European conference on computer vision (ECCV)}, pages 367--383, 2018.

\bibitem{long2018textsnake}
Shangbang Long, Jiaqiang Ruan, Wenjie Zhang, Xin He, Wenhao Wu, and Cong Yao.
\newblock Textsnake: A flexible representation for detecting text of arbitrary shapes.
\newblock In {\em Proceedings of the European conference on computer vision (ECCV)}, pages 20--36, 2018.

\bibitem{vadlamudi2023seamformer}
Niharika Vadlamudi, Rahul Krishna, and Ravi~Kiran Sarvadevabhatla.
\newblock Seamformer: high precision text line segmentation for handwritten documents.
\newblock In {\em International Conference on Document Analysis and Recognition}, pages 313--331. Springer, 2023.

\bibitem{monnier2020docextractor}
Tom Monnier and Mathieu Aubry.
\newblock docextractor: An off-the-shelf historical document element extraction.
\newblock In {\em 2020 17th International Conference on Frontiers in Handwriting Recognition (ICFHR)}, pages 91--96. IEEE, 2020.

\bibitem{rabaev2025recent}
Irina Rabaev and Marina Litvak.
\newblock Recent advances in text line segmentation and baseline detection in historical document images: a systematic review.
\newblock {\em IJDAR}, pages 1--37, 2025.

\bibitem{zottin2025icdar}
Silvia Zottin, Axel De~Nardin, Giuseppe Branca, Claudio Piciarelli, and Gian~Luca Foresti.
\newblock Icdar 2025 competition on few-shot text line segmentation of ancient handwritten documents (fest).
\newblock In {\em ICDAR}, pages 586--602. Springer, 2025.

\bibitem{sterzinger2025few}
Rafael Sterzinger, Tingyu Lin, and Robert Sablatnig.
\newblock Few-shot connectivity-aware text line segmentation in historical documents.
\newblock In {\em ACPR}, pages 100--114. Springer, 2025.

\bibitem{hu2024seghist}
Xingjian Hu, Baole Wei, Liangcai Gao, and Jun Wang.
\newblock Seghist: A general segmentation-based framework for chinese historical document text line detection.
\newblock In {\em ICDAR}, pages 391--410. Springer, 2024.

\bibitem{agrawal2025linetr}
Vaibhav Agrawal, Niharika Vadlamudi, Muhammad Waseem, Amal Joseph, Sreenya Chitluri, and Ravi~Kiran Sarvadevabhatla.
\newblock Linetr: Unified text line segmentation for challenging palm leaf manuscripts.
\newblock In {\em ICPR}, pages 217--233. Springer, 2025.

\bibitem{chincholikar2025case}
Kartik Chincholikar, Shagun Dwivedi, Kaushik Gopalan, and Tarinee Awasthi.
\newblock A case study of handwritten text recognition from pre-colonial era sanskrit manuscripts.
\newblock In {\em Computational Sanskrit and Digital Humanities-World Sanskrit Conference 2025}, pages 52--69, 2025.

\bibitem{xing2019convolutional}
Linjie Xing, Zhi Tian, Weilin Huang, and Matthew~R Scott.
\newblock Convolutional character networks.
\newblock In {\em Proceedings of the IEEE/CVF international conference on computer vision}, pages 9126--9136, 2019.

\bibitem{baek2019character}
Youngmin Baek, Bado Lee, Dongyoon Han, Sangdoo Yun, and Hwalsuk Lee.
\newblock Character region awareness for text detection.
\newblock In {\em Proceedings of the IEEE/CVF conference on computer vision and pattern recognition}, pages 9365--9374, 2019.

\bibitem{baek2020cleval}
Youngmin Baek, Daehyun Nam, Sungrae Park, Junyeop Lee, Seung Shin, Jeonghun Baek, Chae~Young Lee, and Hwalsuk Lee.
\newblock Cleval: Character-level evaluation for text detection and recognition tasks.
\newblock In {\em Proceedings of the IEEE/CVF Conference on Computer Vision and Pattern Recognition Workshops}, pages 564--565, 2020.

\bibitem{kumar2024review}
Vishal Kumar.
\newblock Review of computational epigraphy.
\newblock {\em arXiv preprint arXiv:2406.06570}, 2024.

\bibitem{howe2025character}
Nicholas~R Howe, Feiran Chang, Isabella Falbo, Tajhini Brown, and Aaron Hershkowitz.
\newblock Character recognition for greek squeezes.
\newblock {\em IJDAR}, pages 1--12, 2025.

\bibitem{bhuvaneswari2024enhancing}
S~Bhuvaneswari and K~Kathiravan.
\newblock Enhancing epigraphy: a deep learning approach to recognize and analyze tamil ancient inscriptions.
\newblock {\em Neural Computing and Applications}, 36(31):19839--19861, 2024.

\bibitem{sekharan2025veda}
Sindhu~Chandra Sekharan, Taruni Mamidipaka, Yoga Sreedhar~Reddy Kakanuru, Summia Parveen, and S~Saradha.
\newblock Veda: Visual extraction and decryption of ancient scripts.
\newblock In {\em International Conference on Smart Trends for Information Technology and Computer Communications}, pages 345--355. Springer, 2025.

\bibitem{agrawal2024optical}
Yash Agrawal, Srinidhi Balasubramanian, Rahul Meena, Rohail Alam, Himanshu Malviya, et~al.
\newblock Optical character recognition using convolutional neural networks for ashokan brahmi inscriptions.
\newblock {\em arXiv preprint arXiv:2501.01981}, 2024.

\bibitem{mohammed2024detection}
Hussein Mohammed and Mahdi Jampour.
\newblock From detection to modelling: An end-to-end paleographic system for analysing historical handwriting styles.
\newblock In {\em International Workshop on Document Analysis Systems}, pages 363--376. Springer, 2024.

\bibitem{zhen2024oracle}
Qianqian Zhen, Liang Wu, and Guoying Liu.
\newblock An oracle bone inscriptions detection algorithm based on improved yolov8.
\newblock {\em Algorithms}, 17(5):174, 2024.

\bibitem{tao2025clustering}
Ye~Tao, Xinran Fu, Honglin Pang, Xi~Yang, and Chuntao Li.
\newblock Clustering-based feature representation learning for oracle bone inscriptions detection.
\newblock {\em npj Heritage Science}, 13(1):296, 2025.

\bibitem{fu2024detecting}
Xinran Fu, Rixin Zhou, Xi~Yang, and Chuntao Li.
\newblock Detecting oracle bone inscriptions via pseudo-category labels.
\newblock {\em Heritage Science}, 12(1), 2024.

\bibitem{qi2025ancientglyphnet}
Hengnian Qi, Hao Yang, Zhaojiang Wang, Jiabin Ye, Qiuyi Xin, Chu Zhang, and Qing Lang.
\newblock Ancientglyphnet: an advanced deep learning framework for detecting ancient chinese characters in complex scene.
\newblock {\em Artificial Intelligence Review}, 58(3):88, 2025.

\bibitem{preethi2023region}
Padmaprabha Preethi and Hosahalli~Ramappa Mamatha.
\newblock Region-based convolutional neural network for segmenting text in epigraphical images.
\newblock In {\em Artificial Intelligence and applications}, volume~1, pages 103--111, 2023.

\bibitem{guo2023applications}
An~Guo, Zhan Zhang, Feng Gao, Haichao Du, Xiaokui Liu, and Bang Li.
\newblock Applications of convolutional neural networks to extracting oracle bone inscriptions from three-dimensional models.
\newblock {\em Symmetry}, 15(8):1575, 2023.

\bibitem{zheng2024ancient}
Yi~Zheng, Yi~Chen, Xianbo Wang, Donglian Qi, and Yunfeng Yan.
\newblock Ancient chinese character recognition with improved swin-transformer and flexible data enhancement strategies.
\newblock {\em Sensors}, 24(7):2182, 2024.

\bibitem{murugan2025gan}
Balasubramanian Murugan and P~Visalakshi.
\newblock Gan augmented hybrid transformer network (ghtnet) for ancient tamil stone inscription recognition.
\newblock {\em npj Heritage Science}, 13(1):604, 2025.

\bibitem{shen2025unitext}
Lu~Shen, Zewei Wu, Xiaoyuan Huang, Boliang Zhang, Su-Kit Tang, Jorge Henriques, and Silvia Mirri.
\newblock Unitext: A unified framework for chinese text detection, recognition, and restoration in ancient document and inscription images.
\newblock {\em Applied Sciences}, 15(14):7662, 2025.

\bibitem{diao2025ancient}
Xiaolei Diao, Rite Bo, Yanling Xiao, Lida Shi, Zhihan Zhou, Hao Xu, Chuntao Li, Xiongfeng Tang, Massimo Poesio, C{\'e}dric~M John, et~al.
\newblock Ancient script image recognition and processing: A review.
\newblock {\em arXiv preprint arXiv:2506.19208}, 2025.

\bibitem{carion2020end}
Nicolas Carion, Francisco Massa, Gabriel Synnaeve, Nicolas Usunier, Alexander Kirillov, and Sergey Zagoruyko.
\newblock End-to-end object detection with transformers.
\newblock In {\em ECCV}, pages 213--229. Springer, 2020.

\bibitem{ravi2024sam}
Nikhila Ravi, Valentin Gabeur, Yuan-Ting Hu, Ronghang Hu, Chaitanya Ryali, Tengyu Ma, Haitham Khedr, Roman R{\"a}dle, Chloe Rolland, Laura Gustafson, et~al.
\newblock Sam 2: Segment anything in images and videos.
\newblock {\em arXiv preprint arXiv:2408.00714}, 2024.

\bibitem{yolov8_ultralytics}
Glenn Jocher, Ayush Chaurasia, and Jing Qiu.
\newblock {Ultralytics YOLOv8}.
\newblock \url{https://github.com/ultralytics/ultralytics}, 2025.
\newblock Version 8.0.0, Licensed under AGPL-3.0.

\bibitem{yolo11_ultralytics}
Glenn Jocher and Jing Qiu.
\newblock {Ultralytics YOLO11}.
\newblock \url{https://github.com/ultralytics/ultralytics}, 2025.
\newblock Version 11.0.0, license AGPL-3.0.

\bibitem{tian2025yolov12}
Yunjie Tian, Qixiang Ye, and David Doermann.
\newblock Yolov12: Attention-centric real-time object detectors.
\newblock {\em arXiv preprint arXiv:2502.12524}, 2025.

\bibitem{dosovitskiy2020image}
Alexey Dosovitskiy.
\newblock An image is worth 16x16 words: Transformers for image recognition at scale.
\newblock {\em arXiv preprint arXiv:2010.11929}, 2020.

\end{thebibliography}
%
%
%




\end{document}